\documentclass[conference]{IEEEtran}

\usepackage{cite}

\ifCLASSINFOpdf
  \usepackage[pdftex]{graphicx}
  \graphicspath{{./images/}}
  \DeclareGraphicsExtensions{.pdf,.jpeg,.png}
\else
  \usepackage[dvips]{graphicx}
  \graphicspath{{./images/}}
  \DeclareGraphicsExtensions{.eps}
\fi

\usepackage{amsmath}

\usepackage{url}

\usepackage{enumitem}
\usepackage{subfig}

\usepackage{hyperref}

\newcommand{\currentversion}{0.9}
\newcommand{\numberofsymbols}{91255}
\newcommand{\numberofrelationships}{82261}
\newcommand{\numberofnotes}{23352}
\newcommand{\numberofclasses}{158}
\newcommand{\numberofclassesused}{107}

\begin{document}

\title{In Search of a Dataset for \\ Handwritten Optical Music Recognition:\\ Introducing MUSCIMA++}

\author{\IEEEauthorblockN{Jan Haji\v{c} jr.}
\IEEEauthorblockA{Institute of Formal and Applied Linguistics
\\Faculty of Mathematics and Physics\\
Charles University\\
Email: hajicj@ufal.mff.cuni.cz}
\and
\IEEEauthorblockN{Pavel Pecina}
\IEEEauthorblockA{Institute of Formal and Applied Linguistics
\\Faculty of Mathematics and Physics\\
Charles University\\
Email: pecina@ufal.mff.cuni.cz}
}

\maketitle

\begin{abstract}
Optical Music Recognition (OMR) has long been without an adequate
dataset and ground truth for evaluating OMR systems, which has been
a major problem for establishing a state of the art in the field.
Furthermore, machine learning methods require training data.
We analyze how the OMR processing pipeline can be
expressed in terms of gradually more complex ground truth,
and based on this analysis we present the MUSCIMA++ dataset
of handwritten music notation that addresses musical symbol recognition 
and notation reconstruction.
The MUSCIMA++ dataset v.\currentversion~
consists of 140 pages of handwritten music, 
with~\numberofsymbols~manually 
annotated notation symbols and~\numberofrelationships~explicitly 
marked relationships between symbol pairs.
The dataset allows training and evaluating models for symbol classification,
symbol localization, and notation graph assembly, both in isolation
and jointly.
Open-source tools are provided for manipulating the dataset,
visualizing the data and annotating further, and the dataset
itself is made available under an open license.
\end{abstract}

\section{Introduction: what dataset?}
\label{sec:introduction}

Optical Music Recognition (OMR) is a field of document analysis
that aims to automatically read musical scores.
Music notation encodes the musical information in a graphical form;
OMR backtracks through this process to extract the musical information
from this graphical representation.

OMR can perhaps be likened to OCR for the music notation writing system;
however, it is more difficult
\cite{Bainbridge2001}, and remains an open problem
\cite{Rebelo2012, Novotny2015}. Common western music
notation (CWMN\footnote{We assume the reader is familiar
with this style of music notation and its terminology. In case
a refresher is needed, we recommend the excellent chapter 2
of ``Music Notation by Computer'' \cite{Byrd1984},
by Donald Byrd.})
is an intricate writing system, where both the vertical
and horizontal dimensions are salient and used to resolve symbol
ambiguity.
In terms of graphical complexity, the biggest issues are caused by
 overlapping symbols (including stafflines) \cite{Bainbridge1997b}
and composite symbol constructions.
In handwritten music, the variability of handwriting leads
 to a lack of reliable topological properties overall.
And in polyphonic music, there
are multiple sequences written, in a sense, ``over'' each other
 -- as opposed to OCR, where the ordering of the symbols is linear.

Moreover, the objective of OMR is
more ambitious than OCR: recovering not just the locations and classification of
musical symbols, but also ``the music'': pitch and duration information
of individual notes. This introduces long-distance relationships: the interpretation
of one ``letter'' of music notation may change based on notation events
some distance away. These intricacies of music notation
has been thoroughly discussed since early attempts at OMR, notably
by Byrd \cite{Byrd1984, Byrd2015}.

One of the most persistent problems that have hindered OMR progress
is the {\bf lack of datasets}. These are necessary to provide ground truth
for evaluating OMR systems \cite{Bainbridge2001, Droettboom2004, Padilla2014, Chanda2014, Shi2015, Byrd2015, HajicJr2016, Baro2016}.
This is frustrating, because the many proposed recognition systems
cannot be compared.
Furthermore, especially for handwritten notation, statistical
machine learning methods have often been used that require
training data for parameter estimation
\cite{Stuckelberg1999, Rebelo2011, Rebelo2011a, CalvoZaragoza2014, Wen2015}.  

For printed music notation, the lack of datasets can be bypassed
by rendering sheet music images from synthesized representations such as
LilyPond\footnote{\url{http://www.lilypond.org}} 
or MEI,\footnote{\url{http://www.music-encoding.org}}  
capturing intermediate steps,
and using data augmentation techniques to simulate
desired deformations and lighting conditions.\footnote{We are aware of one such ongoing effort for the LilyPond format: \url{http://lilypond.1069038.n5.nabble.com/Extracting-symbol-location-td194857.html}}
However, for handwritten music, no satisfactory
synthetic data generator exists so far, and an extensive (and expensive)
annotation effort cannot be avoided. Even if a synthesis
system were to be implemented, it needs to be evaluated against
actual handwritten data -- so, at least some manual annotation is needed.
Therefore, in order to best utilize the resources available
for creating a dataset, we decided to create a dataset of {\bf handwritten notation}.

%
%
%

We use the term {\em dataset} in the following sense:
$\mathcal{D} = \langle \left( x_i, y_i \right) \: \forall i = 1 \dots n \rangle$.
Given a set of inputs $x_i$ (in our case, images of sheet music), the dataset
records the desired outputs $y_i$ -- ground truth. The quality of OMR systems can
then be measured by how closely they approximate the ground truth, although
defining this approximation for the variety of representations of music
is very much an open problem
\cite{Droettboom2004, Bellini2007, Szwoch2008, Rebelo2012, Byrd2015, HajicJr2016}.

In order to build a dataset of handwritten music, we need to decide on two issues:

\begin{itemize}
\item What sheet music do we choose as data points?
\item What should the desired output $y_i$ be for an image $x_i$?
\end{itemize}

The music scores in the dataset should 
{\bf cover the known ``dimensions of difficulty''},
to allow for assessing OMR systems with respect to increasingly complex inputs.
There are two main categories of challenges that make OMR less or more
difficult: image-related difficulty,
and complexity of the music that we are trying to recognize \cite{Byrd2015}.
While image-related difficulties such as uneven lighting or deformations can be
simulated, the complexity of notation is not a ``knob to turn'' and needs
to be explicitly relfected by the choice of included music. The dataset
needs to at least include music of all the four basic
levels of difficulty according to Byrd \& Simonsen \cite{Byrd2015}:
single-staff single-voice, single-staff multi-voice, multi-staff single-voice,
and ``pianoform'' music.

\begin{figure}[!t]
\centering
\subfloat[Input: manuscript image.]{\includegraphics[width=3.0in]{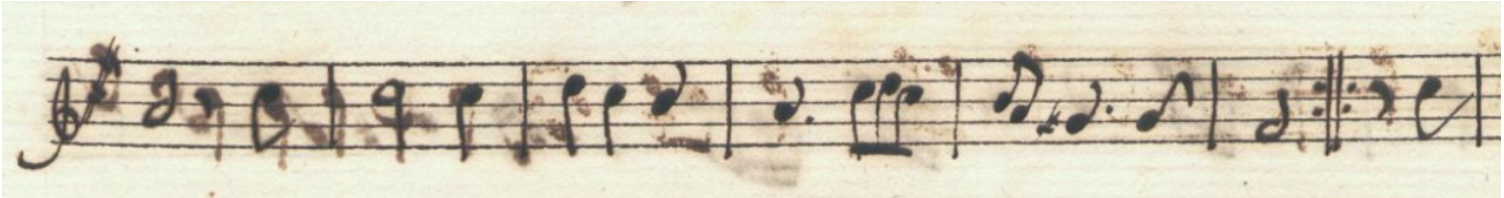}%
\label{fig:goals:input}}
\vfil
\subfloat[Replayable output: pitches, durations, onsets. Time is the horizontal axis, pitch is the vertical axis. This visualization is called a {\em piano roll}.]{\includegraphics[width=3.0in]{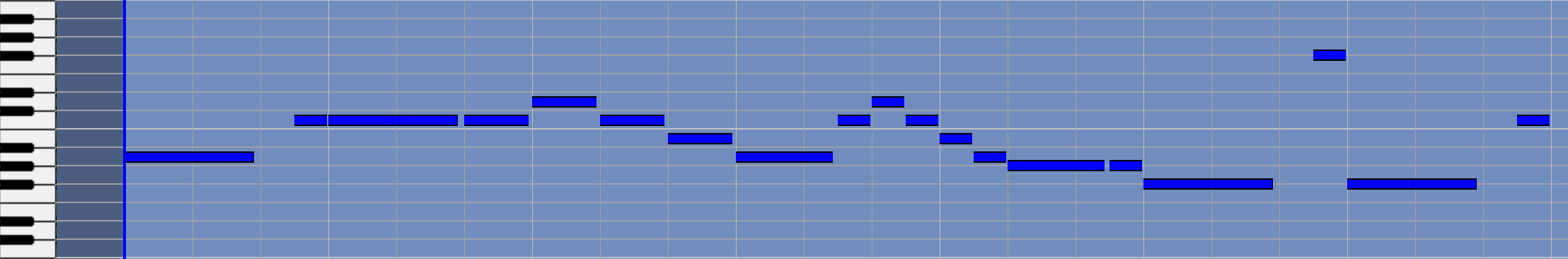}%
\label{fig:goals:replayable}}
\vfil
\subfloat[Reprintable output: re-typesetting.]{\includegraphics[width=3.0in]{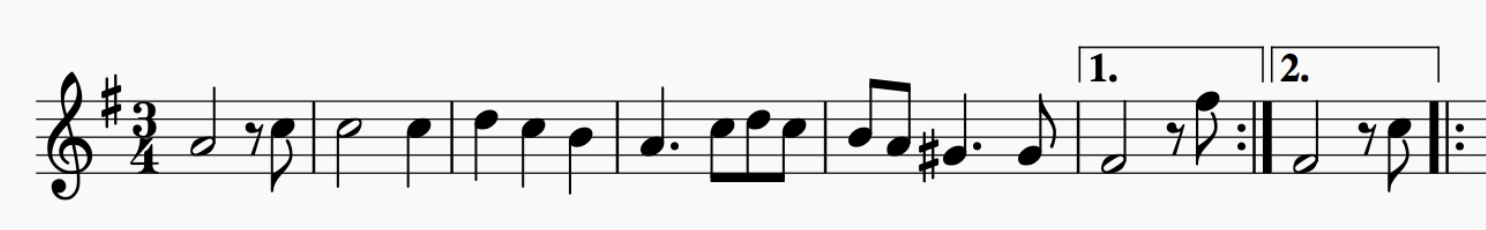}
\label{fig:goals:reprintable}}
\vfil
\subfloat[Reprintable output: same music expressed differently]{\includegraphics[width=3.0in]{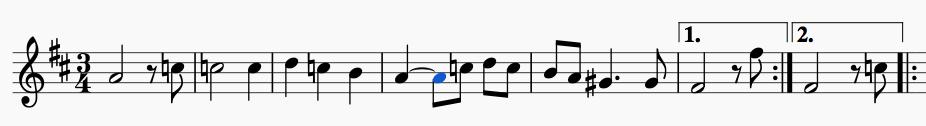}
\label{fig:goals:reprintabledifferent}}
\caption{OMR for replayability and reprintability. The input (a) 
encodes the sequence of pitches,
durations, and onsets (b), which can be expressed in different ways (c, d).}
\label{fig:goals}
\end{figure}

The ground truth must {\bf reflect
what OMR {\em does}.} Miyao and Haralick \cite{Miyao2000}
suggest grouping OMR applications into two broad groups: 
those that require replayability,
and those that need reprintability.
{\bf Replayability} is understood to consist of recovering
pitches and durations of individual notes and organizing them
as a time series by note onset. (For the purposes of this article,
``musical sequence'' will refer to this minimum replayable data.) 
{\bf Reprintability} means the ability to take OMR results as the input
to music typesetting software and obtain a result that is equivalent
to the input. Reprintability implies replayability, but not vice versa, as
one musical sequence can be encoded by different musical scores
(see fig. \ref{fig:goals}).
OMR systems 
are usually pipelines with
four major stages
\cite{Bainbridge2001, Rebelo2012, Hankinson2015Thesis}:

\begin{enumerate}[noitemsep]
\item Image preprocessing: enhancement, binarization, scale normalization;
\item Music symbol recognition: staffline identification and removal, localization and classification of remaining symbols;
\item Musical notation reconstruction: recovering the logical structure of the score;
\item Final representation construction: depending on the output requirements, usually inferring pitch and duration (MusicXML, MEI, MIDI, LilyPond, etc.).
\end{enumerate}

\noindent
While end-to-end OMR that bypasses some sections
of this pipeline is clearly an option (see \cite{Shi2015}),
and might yet make this pipeline obsolete,
there is still a need in the field to compare new systems 
against more orthodox solutions.
Therefore, to design a dataset broadly useful to the OMR 
community, we first need to examine how the stages 
of OMR pipelines  can be expressed in terms of inputs 
and outputs. Then, we can identify what ground truth is needed.

After the fundamental design decisions are made, 
we also need to consider additional factors:

\begin{itemize}
\item Economy, which reduces to: what is the smallest subset of the ground truth that absolutely has to be annotated manually per item $x_i$?
\item Compatibility: can existing results on related datasets be compared directly with results on the new dataset?
\item Intellectual property rights: can the data be released under an open license, such as the Creative Commons family?
\item Ease of use: how difficult is it to take the dataset and run an experiment?
\end{itemize}

\noindent
Once the dataset is designed, we can proceed to build an annotation interface
and annotate the data.

It should also be understood that while the dataset records its ground truth
in some representation, it is not trying to enforce it as {\em the} representation
that should be used for specific experiments. Rather, experiment-specific
output representations (such as pitch sequences for end-to-end experiments
by Shi et al. in \cite{Shi2015}, or indeed a MIDI file)
should be {\em unambiguously derivable} from the dataset.
When defining the ground truth, we are concerned with
{\em what information to record}, not necessarily {\em how} to record it.
However, we want to be sure that we record enough
about the sheet music in question, so that the dataset is useful for a wide
range of purposes. The choice of dataset representation is made to give
some theoretical guarantees on this ``information content''.

\subsection{Contributions and outline}
\label{subsec:introduction:outline}

The main contributions of this work are:

\begin{itemize}[noitemsep]
\item MUSCIMA++\footnote{Standing for MUsic SCore IMAges, credit for abbreviation to \cite{Fornes2012}} -- an extensive dataset 
of handwritten musical symbols,\footnote{Available from: \url{https://ufal.mff.cuni.cz/muscima}}
\item A principled ground truth definition that 
bridges the gap between the graphical expression of music and musical semantics, 
enabling evaluation of multiple OMR sub-tasks up to inferring pitch and duration
both in isolation and jointly;
\item Open-source tools for processing the data, visualizing it and annotating more.
\end{itemize}

The rest of this article is organized into five sections.
In section \ref{sec:groundtruth}, we describe the OMR pipeline
through input-output interfaces, in order to design a good ground truth
for the dataset.

In section \ref{sec:choiceofdata}, we discuss what 
kinds of data should be represented by the dataset -- how to choose
images to annotate (on a budget) in order to maximize impact for OMR?

In section \ref{sec:existingdatasets}, we describe existing datasets
for musical symbol recognition, especially those that are publicly available,
with respect to the requirements formulated in \ref{sec:groundtruth}
and \ref{sec:choiceofdata}. While there are no datasets that would satisfy
both, we have concluded that the CVC-MUSCIMA dataset of Forn\'{e}s et al.
\cite{Fornes2012} forms a sound basis.


In section \ref{sec:data}, we then describe 
the current MUSCIMA++ version~\currentversion~in practical terms:
what images we chose to annotate, the data acquisition process, 
tools associated with the dataset,
and we discuss its strengths and limitations.

Finally, in section \ref{sec:conclusion}, we briefly summarize the work,
discuss baselines and evaluation procedures for using the dataset
in experiments, and sketch a roadmap for future versions of the dataset.


\section{Ground truth of musical symbols}
\label{sec:groundtruth}

The ground truth over a dataset is the desired output
of a perfect\footnote{More accurately, as perfect as possible, 
given how the ground truth was acquired. 
See \ref{subsec:data:annotation}.} system solving a task.
In order to design the ground truth for the dataset,
we need to understand which task is it designed to simulate.
How can the OMR sub-tasks be expressed
terms of inputs and outputs?

\subsection{Interfaces in the OMR pipeline}
\label{subsec:groundtruth:stages}

The {\bf image preprocessing} stage is mostly practical: by making the image
conform to some assumptions (such as: stems are straight, attempted by de-skewing),
the OMR system has less complexity to deal with down the road,
while very little to no information is lost. 
The problems that this stage needs to handle are mostly related to document
quality (degradation, stains, especially bleedthrough) and imperfections in
the imaging process (e.g., uneven lighting, deformations of the paper; 
with mobile phone cameras, limited depth-of-field may lead to 
out-of-focus segments of the image) \cite{Byrd2015}.
The most important
problem for OMR in this stage is binarization \cite{Rebelo2012}:
sorting out which pixels
belong to the background, and which actually make up the notation.
There is evidence that sheet music has some specifics
in this respect \cite{JohnAshley2008}, and there have been attempts to tackle binarization for OMR specifically \cite{JaeMyeongYoo2008, Pinto2010, Pinto2011}.
On the other hand, other authors have attempted to bypass binarization,
especially before staffline detection \cite{Rebelo2013a, CalvoZaragoza2016d}.

The input of {\bf music symbol recognition} is a ``cleaned'' and usually
binarized image. The output of this stage is a list of musical symbols
recording their locations on the page, and their types (e.g., c-clef, beam, sharp).
Usually, there are three sub-tasks: staffline identification and removal,
symbol localization (in binary images, synonymous with foreground segmentation),
and symbol classification \cite{Rebelo2012}.

Stafflines are usually handled first. 
Removing them then greatly simplifies the foreground,
and in turn the remainder of the task, as it will make
connected components a useful (if imperfect)
heuristic for pruning the search space of possible segmentations
\cite{Fujinaga1996, Rebelo2012Thesis}. Staffline detection and removal
is a large topic in OMR since its inception
\cite{Pruslin1966, Prerau1971, Fujinaga1988},
and it is the only one where a competition
was successfully organized, by Forn\'{e}s et al. \cite{fornes2011icdar}.
Because errors during staff removal make further recognition complicated,
especially by breaking symbols into multiple connected components
with over-eager removal algorithms, some authors skip this stage:
Sheridan and George instead add extra stafflines to annul differences
between notes on stafflines and between stafflines \cite{Sheridan2004}, 
Pugin interprets the stafflines in a symbol's bounding box to be part
of that symbol for the purposes of recognition \cite{Pugin2006}.

\begin{figure}[!t]
\centering
\includegraphics[width=3.0in]{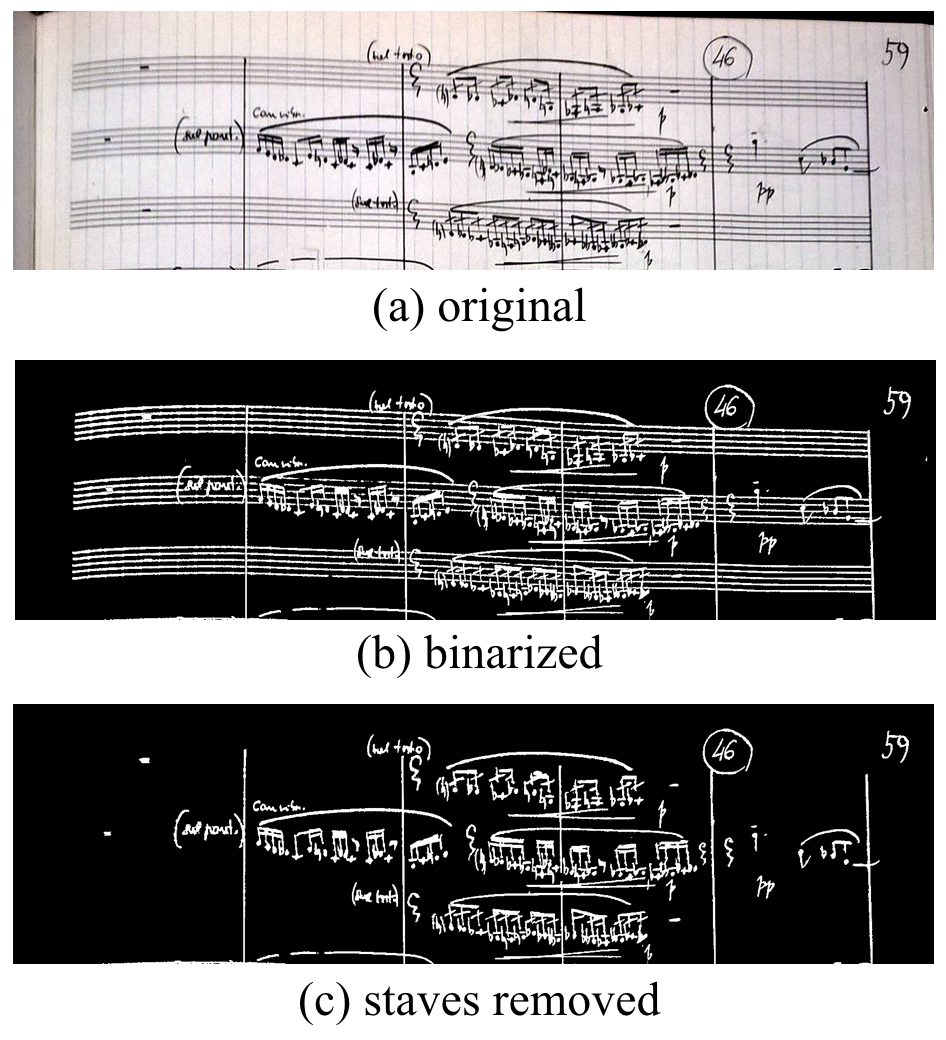}
\caption{OMR pipeline from the original image through image processing, binarization, and staff removal. While staff removal is technically part of symbol recognition, as stafflines are symbols as well, it is considered practical to think of the image after staff removal as the input for recognition of other symbols.}
\label{fig:origtostaffremoval}
\end{figure}


Next, the page is segmented into musical symbols, and
the segments are classified by symbol class. While classification of musical
symbols has produced near-perfect scores for both printed and handwritten
musical symbols \cite{Rebelo2012Thesis, Wen2016}, segmentation of handwritten scores remains elusive, as
most segmentation approaches such as projections 
\cite{Fujinaga1988, Fujinaga1996, Bellini2001}
rely on topological constraints that do not necessarily
hold in handwritten music. Morphological skeletons have been proposed 
instead \cite{Ng1999, Luth2002} as a basis for handwritten OMR.

This stage is where the first non-trivial ground truth design decisions need to be
taken: the ``alphabet'' of elementary music notation symbols must be defined.
Some OMR researchers decompose notation into individual primitives
(notehead, stem, flag)
\cite{Couasnon1994, Bainbridge1997, Bellini2001, Bainbridge2003, Fornes2005Thesis},
while others retain the ``note'' as an elementary visual object.
Beamed groups are decomposed into the beam(s) and the remaining
notehead+stem ``quarter-like notes'' 
\cite{Rebelo2010, Rebelo2012Thesis, Pham2015}, 
or not included \cite{CalvoZaragoza2014, Chanda2014}.

There are some datasets available for symbol recognition, although 
except for staff removal, they
do not yet respond to the needs of the OMR community, especially
since most machine learning efforts have been directed to this stage;
see section \ref{sec:existingdatasets}. 

In turn, the list of locations and classes of symbols on the page is the input
to the {\bf music notation reconstruction} stage. The outputs are
more complex: at this stage, it is necessary to recover the relationships
among the individual musical symbols, so that from the resulting representation,
the ``musical content'' (most importantly, pitch and duration information --
what to play, and when) can be unambiguously inferred.

There are two important observations to make at stage 3.

First, with respect to pitch and duration, the rules of music notation finally
give us something straightforward: {\em there is a 1:1 relationship between 
a notehead notation primitive and a note musical object}, 
of which pitch and duration are properties.\footnote{Ornaments, such as 
trills or glissandos, do not really count as notes in common-practice 
music understanding. They are understood as ``complications'' of a note
or a note pair.} 
The other symbols that relate to a notehead,
such as stems, stafflines, beams, accidentals (be they attached
to a note, or part of a key signature), or clefs, 
inform the reader's decision to assign the pitch and duration.

Second, a notable property of this stage is that 
once inter-symbol relationships are fully recovered (including
precedence, simultaneity and high-level relationships between staves),
{\em symbol positions cease to be informative:} they serve
primarily as features that help music readers infer these relationships.
If we wanted to, we could forget the input image after stage 3.

However, it is unclear how these relationships
should be defined. For instance: should there be relationships of multiple types?
Is the relationship between a notehead and a stem different in principle than between
a notehead and an associated accidental? Or the notehead and the key signature?
If the note is written as an {\tt f}, and the key is D major (two sharps,
on {\tt c} and on {\tt f}): does the note relate to the entire key signature,
or just to the accidental that directly modifies it? What about the relationship
between barlines and staves?

Instead of a notation reconstruction stage, 
other authors define two levels of symbols: {\em low-level} primitives
that cannot by themselves express musical semantics, and {\em high-level} symbols
that already do have some semantic interpretation \cite{Bellini2007, Byrd2015}.
For instance, the letter {\tt p} is just a letter from the low-level point of view, 
but a dynamics sign from the high-level perspective. This is a distinction
made when discussing evaluation, in an attempt to tie errors in semantics
to units that can be counted.

We view this approach as complementary: 
the high-level symbols can also belong to the symbol set
of stage 2, and the two levels of description can be explicitly linked.
The fact that correctly interpreting whether the {\tt p} is a dynamics sign,
or part of a text (e.g., {\tt presto}) requires knowing the positions
and classes of other symbols, simply hints that it may be a good idea
to solve stages 2 and 3 jointly.

Naturally, the set of relationships over a list of elementary music notation
graphical elements (symbols) can be represented by a graph,
possibly directed and requiring labeled edges. The list of symbols
from the previous step can be re-purposed as a list of vertices of the graph,
with the symbol classes and locations being the vertex attributes.
We are not aware of a publicly available dataset that addresses this level,
although at least for handwritten music, this is also a non-trivial step.
Graph-based assembly of music notation primitives has been used
explicitly e.g. by \cite{Reed1996, Chen2015},
and grammar-based approaches
(e.g., \cite{Fujinaga1988, Couasnon1994, Bainbridge2003, Szwoch2007
}
) lend themselves to a graph
representation as well, by recording the parse tree(s).

\begin{figure*}[!t]
\centering
\subfloat[Notation symbols: noteheads, stems, beams, ledger lines, a duration dot, slur, and ornament sign; part of a barline on the lower right. Vertices of the notation graph.]{\includegraphics[width=2.5in]{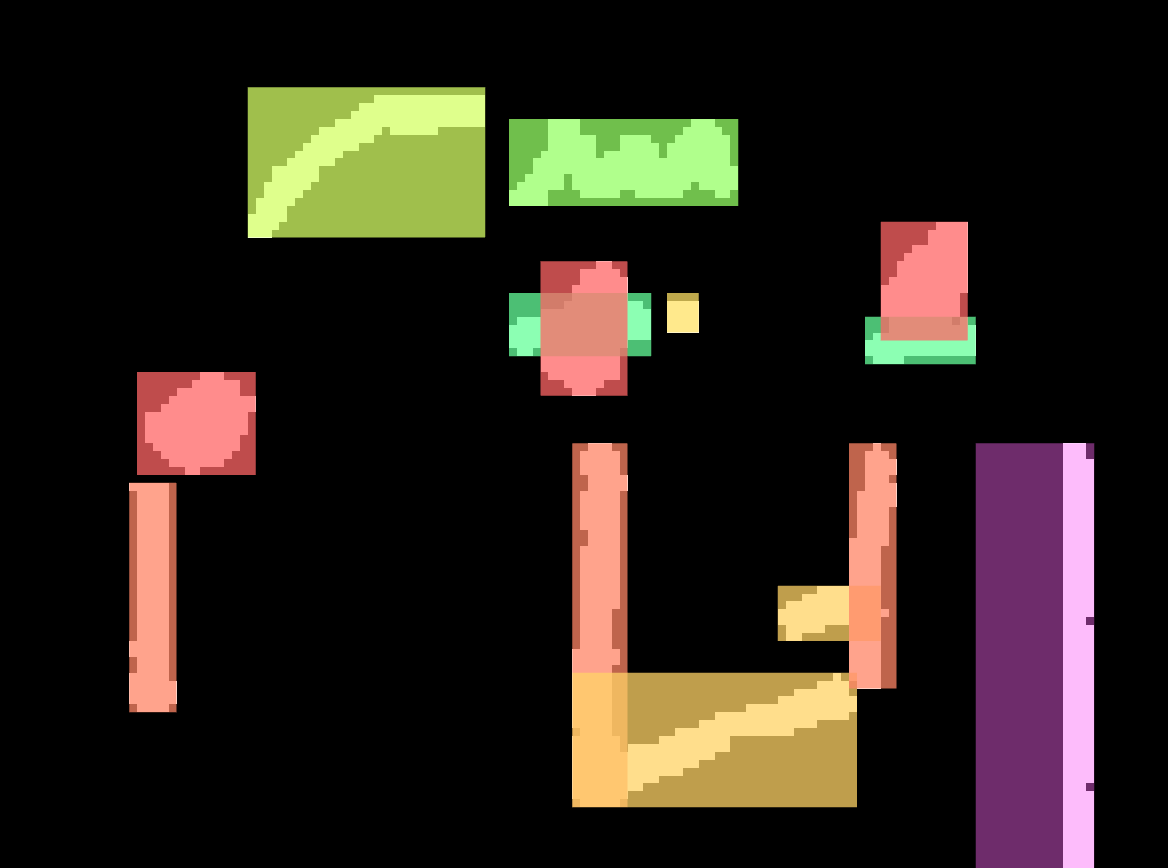}%
\label{fig:notationgraph:vertices}}
\hfil
\subfloat[Notation graph, highlighting noteheads as ``roots'' of subtrees. Noteheads share the beam and slur symbols.]{\includegraphics[width=2.5in]{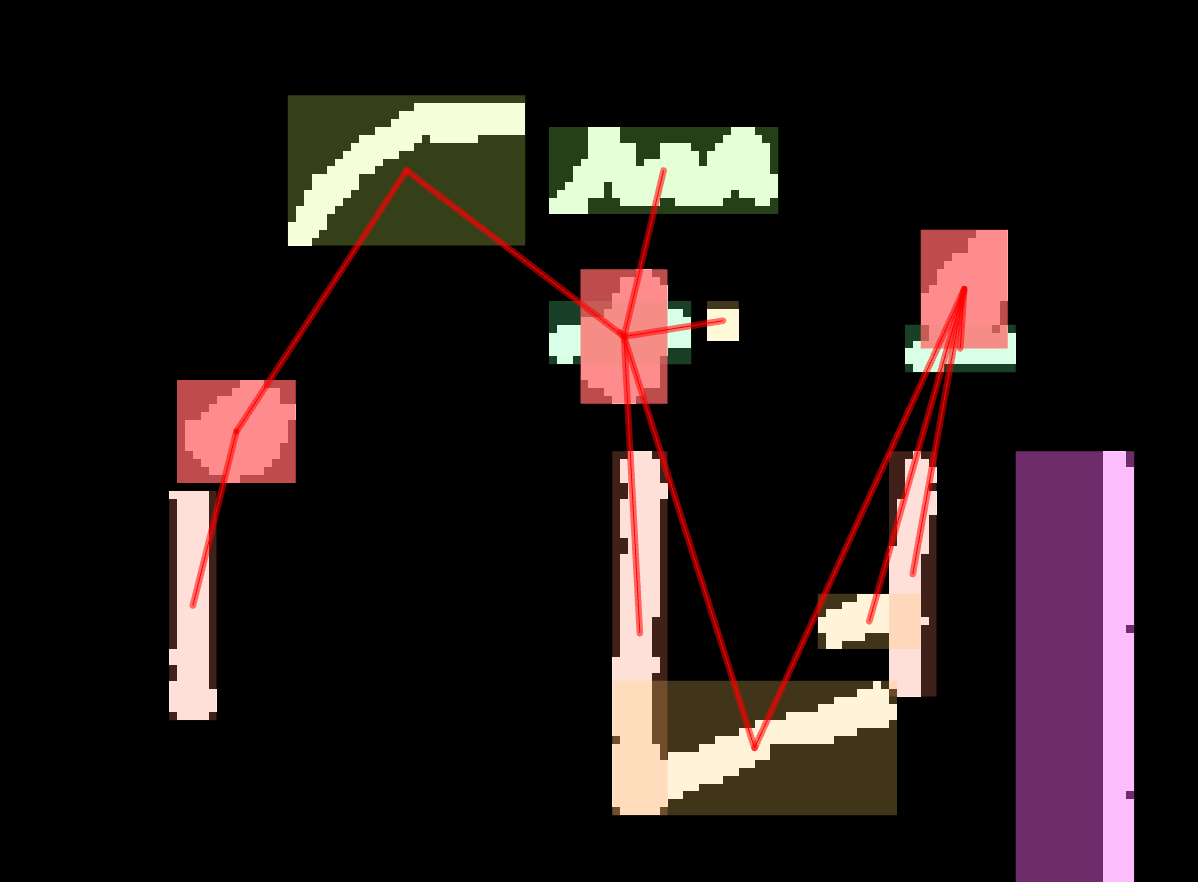}%
\label{fig:notationgraph:edges}}
\caption{Visualizing the list of symbols and the notation graph over staff removal output. The notation graph in (b) allows unambiguously inferring pitch and duration.}
\label{fig:notationgraph}
\end{figure*}

Finally, the chosen output representation of the music notation reconstruction
step must be a good input for the next stage: {\bf encoding the information in
the desired output format.}
There is a plethora of music encoding formats:
from the text-based formats such as 
DARMS,\footnote{\url{http://www.ccarh.org/publications/books/beyondmidi/online/darms/} -- under the name Ford-Columbia music representation,
used as output already in the DO-RE-MI system of Prerau, 1971 \cite{Prerau1971}.)} 
{\tt **kern},\footnote{\url{http://www.music-cog.ohio-state.edu/Humdrum/representations/kern.html}}
LilyPond,
ABC,\footnote{\url{http://abcnotation.com/}}
over NIFF,\footnote{\url{http://www.music-notation.info/en/formats/NIFF.html}} 
MIDI,\footnote{\url{https://www.midi.org/}} 
to XML-based formats MusicXML\footnote{\url{http://www.musicxml.com/}} 
and MEI.
The individual formats are each suitable for a different purpose: for instance,
MIDI is most useful for interfacing different electronic audio devices, 
MEI is great for editorial work,
LilyPond allows for excellent control of music engraving.
Many of these have associated software tools that enable rendering the encoded
music as a standard musical score, although some -- notably MIDI -- do not allow for
a lossless round-trip. Furthermore, evaluating against the more complex formats
is notoriously problematic \cite{Szwoch2008, HajicJr2016}.

Text-based formats are also ripe targets for end-to-end OMR, as they
reduce the output to a single sequence, which enables the application
of OCR, text-spotting, or even image-captioning models. This has been
attempted specifically for OMR by Shi et al. \cite{Shi2015} using
a recurrent-convolutional neural network -- although
with only modest success on a greatly simplified task.
However, even systems that only use stage 4 output 
and do not use stage 2 and 3 output in an intermediate step 
have to consider stage 2 and 3 information implicitly: 
that is simply how music notation conveys meaning.

That being said, stage 4 is -- or, with a good stage 3 output, could be -- 
mostly a technical step. The output of the notation reconstruction stage 
should leave as little ambiguity as possible for this last step to handle. 
In effect, the combination of outputs of the previous 
stages should give a potential user enough information to construct
the desired representation for {\em any} task that may come up and does not
require more input than the original image: after all, the musical
score contains a finite amount of information, and it can be explicitly
represented.\footnote{This does not mean, however, that the score
contains by itself enough information to {\em perform} the music.
That skill requires years of training, experience, familiarity
with tradition, and scholarly expertise, and is not a goal of OMR systems.}

\begin{table}[!t]
\renewcommand{\arraystretch}{1.3}
\caption{OMR Pipeline: Inputs and Outputs}
\label{tab:groundtruth:io}
\centering
\begin{tabular}{c|cc}
{\bf Sub-task} & {\bf Input} & {\bf Output} \\
\hline
\hline
Image Processing & Score image & ``Cleaned'' image \\
Binarization & ``Cleaned'' image & Binary image \\
\hline
Staff ID \& removal & Binary image & Stafflines list \\
Symbol localization & (Staff-less) image & Symbol regions \\
Symbol classification & Symbol regions & Symbol labels \\
\hline
Notation assembly & Symbol regs. \& labels & Notation graph \\
Infer pitch/duration & Notation graph & Pitch/duration attrs.\\
\hline
Output conversion & Notation graph + attrs. & MusicXML, MIDI, ... \\
\end{tabular}
\end{table}

Table \ref{tab:groundtruth:io} summarizes the steps of OMR and their
inputs and outputs.

It is appealing to approach some of these tasks jointly. For instance,
classification and notation primitive assembly inform each other:
a vertical line in the vicinity of an elliptical blob is probably a stem with its notehead,
rather than a barline.\footnote{This is also a good reason for not decomposing notes
into stem and notehead primitives in stage 2, as done by 
Rebelo et al. \cite{Rebelo2010}:
instead of dealing with graphically ambiguous
symbols, define symbols so that they are graphically more distinct.}
Certain output formats allow leaving outputs
of previous stages underspecified: if our application requires only ABC notation,
then we might not really need to recover regions and labels for slurs
or articulation marks. However, the ``holy grail'' of handwritten OMR, 
transcribing manuscripts to a more readable (reprintable) and shareable form,
requires retaining as much information from the original score as possible,
and so there is a need for a benchmark that does {\em not} underspecify.

\subsection{So -- what should the ground truth be?}
\label{subsec:groundtruth:requirements}

%
%
We believe that
there is a strong case for having an OMR dataset with ground truth at stage 3.


The key problems that OMR needs to solve reside in stages 2 and 3.
We argue that we can define stage 3 output as the point where
{\em all ambiguity in the written score
is resolved.}\footnote{There is also ambiguity that cannot be resolved
using solely the written score: for instance, how fast is {\em Adagio}?
This may not seem too relevant, until we attempt searching audio databases
using sheet music queries, or vice versa.} 
That implies that at the end of stage 3, 
all it takes to create the desired representation
in stage 4 is to write a deterministic format conversion from whatever
the stage 3 output representation is to the desired output format (which can
nevertheless still be a very complex piece of software).
Once ambiguity is resolved in stage 3, 
the case can be made that OMR as a field of research has (mostly) done its job. 

At the same time, we cannot call the science finished earlier than that.
The notation graph of stage 3 is the first point
where we have mined all the information from the input image, 
and therefore  we are properly ``free'' to forget about it, if we need to.
In a sense, the notation graph is at the same time a maximal compression
of the musical score, and agnostic with respect to
software used for rendering the score (or analyzing it further).

Of course, because OMR solutions will not be
perfect anytime soon, there is still the need to address stage 4, in order
to optimize the tradeoffs in stages 2 and 3 for the specific purpose that
drives the adoption of a stage 4 output. For instance, 
a query-by-humming system for sheet music search
might opt for an OMR component that is not very good at recognizing barlines
or even durations, but has very good pitch extraction performance.
However, even such partial OMR will still need to be evaluated 
with respect to the individual sub-tasks of stages 2 and 3, 
even though individual components of overall performance
may be weighed differently.
Moreover, even on the way from sheet
music to just pitches, we need a large subset of stage 3 output anyway.
Again, one can hardly err on the side of explicitness when designing
the ground truth.

It seems that at least as long as OMR is decomposed
into the stages described in the previous section, there is need
for a dataset providing ground truth for the various subtasks
all the way to stage 3. We have discussed how to express
these subtasks in terms of inputs and outputs. At the end,
our analysis shows that
{\bf a good ground truth for OMR should contain:}

\begin{itemize}[noitemsep]
\item {\bf The positions and classes of music notation symbols,}
\item {\bf The relationships between these symbols.}
\end{itemize}

\noindent
Stafflines and staves are counted among music notation symbols.
%
%
%
The following design decisions need to be made:

\begin{itemize}[noitemsep]
\item Defining a vocabulary of music notation symbols,
\item Defining the attributes associated with a symbol,
\item Defining what relationships the symbols can form.
\end{itemize}

There are still multiple ways of defining the stage 3 output graph 
in a way that satisfies the disambiguation requirement. 
Consider an isolated 8th note. 
Should edges expressing attachment lead from the notehead to the stem
and from the stem to the flag, or from the notehead to the flag directly?
Should we perhaps express notation as a constituency tree and instead
have an overlay ``8th note'' high-level symbol that has edges leading
to its  component notation primitives?
However, as much as possible, these should be technical choices: 
whichever option is selected,
it should be possible to write an unambiguous script to convert it
to any of these possible representations.
This space of options equivalent in their information content
is the space in which we apply the secondary criteria of cost-efficient annotation,
user-friendliness, etc.

We now have a good understanding of what the ground truth should entail.
The second major question is the choice of data.

\section{Choice of data}
\label{sec:choiceofdata}

What musical scores should be part of an OMR dataset? 

The dataset should enable evaluating handwritten OMR with respect
to the ``axes of difficulty'' of OMR. It should be possible, based on the dataset,
to judge -- at least to some extent -- how systems perform in the face
of the various challenges within OMR.
However, annotating
musical scores at the level required by OMR, as analyzed in section
\ref{sec:groundtruth}, is expensive, and the pages need to be chosen
with care to maximize impact for the OMR community. We are therefore
keen to focus our efforts on representing in the dataset only those variations
on the spectrum of OMR challenges that have to be provided through human
labor and cannot be synthesized.

What is this ``challenge space'' of OMR?
In their state-of-the-art analysis of the difficulties of OMR,
Byrd and Simonsen \cite{Byrd2015} identify three axes along which 
musical score images become less or more challenging inputs for an OMR system:

\begin{itemize}[noitemsep]
\item Notation complexity
\item Image quality
\item Tightness of spacing
\end{itemize}

\noindent
Not discussed in \cite{Byrd2015} is the variability of handwriting styles,
but we argue below it is in fact a generalization of the Tightness of spacing
axis.

The dataset should also contain a wide variety of musical {\em symbols}, including
less frequent items such as tremolos or glissandi, to enable differentiating systems
also according to the breath of their vocabulary.

\subsection{Notation complexity}
\label{subsec:choiceofdata:notationcomplexity}

The axis of notation complexity is structured by \cite{Byrd2015} into four levels:

\begin{enumerate}[noitemsep]
\item Single-staff, monophonic music (one note at a time),
\item Single-staff, multi-voice music (chords or multiple simultaneous voices),
\item Multiple staves, monophonic music on each
\item Multiple staves, ``pianoform'' music.
\end{enumerate}

\noindent
The category of pianoform music is defined as multi-staff, polyphonic, with interaction
between staves, and with no restrictions on how voices appear, disappear,
and interact throughout.

We should also add another level between multi-staff monophonic and pianoform
music: multi-staff music with multi-voice staves, but without notation complexities
specific to the piano, as described in \cite{Byrd2015} appendix C. Many orchestral
scores with {\em divisi} parts\footnote{The instruction {\em divisi} is used 
when the composer intends the players within one orchestral group to split into two
or more groups, each with their own voice.} would fall into this category,
as well as a significant portion of pre-19th century music for keyboard instruments.

Each of these levels brings a new challenge. Level 1 tests an ``OMR minimum'':
the recognition of individual symbols for a single sequence of notes. Level 2
tests the ability to deal with multiple sequences of notes in parallel, so e.g.
rhythmical constraints based on time signatures \cite{Rebelo2013, Padilla2014}
are harder to use (but still applicable \cite{Jin2012}). Level 3 tests
high-level segmentation into systems and staffs; this is arguably easier
than dealing with the polyphony of level 2 \cite{Riba2015},
as the voices on the staves
are not allowed to interact. Our added level 3.5, multi-staff multi-voice,
is just a technical insertion that combines these two difficulties - system segmentation and polyphony. 
Level 4, pianoform music, then presents
the greatest challenge, as piano music has perused the rules of CWMN
to their fullest \cite{Byrd2015} and sometimes beyond.\footnote{The 
first author, Donald Byrd, 
has been maintaining a list of interesting music notation events. See: \url{http://homes.soic.indiana.edu/donbyrd/InterestingMusicNotation.html}}

Can we automatically simulate moving along the axis of notation complexity? 
The answer is: no. While it might be possible with printed music, to some extent,
there is currently no viable system for synthesis of handwritten musical scores.
Therefore, this is an axis of variability that has to be handled through data selection.

\subsection{Image quality}
\label{subsec:choiceofdata:imagequality}

The axis of image quality is discretized in \cite{Byrd2015} into
five grades, with the assumption of obtaining the image using a scanner.
%
We believe that these degradations can be simulated.
Their descriptions essentially provide a how-to: increasing salt-and-pepper
noise, random pertrubations of object borders, and distortions
such as kanungo noise or localized thickening/thinning operations.
While implementing such a simulation is not quite straightforward,
many morphological distortions have already been introduced for
staff removal data \cite{Dalitz2008, Fornes2012}.

Because \cite{Byrd2015} focus is on scanned images, as found e.g. in the IMSLP
database,\footnote{\url{http://www.imslp.org}} the article does not discuss
images taken by a camera, especially a mobile phone, even though
this mode of obtaining images is widespread and convenient.
The difficulties involved in musical score
photos are mostly uneven lighting and page deformation. 

Uneven lighting
is a problem for binarization, and given sufficient training data,
it can quite certainly be defeated by using convolutional neural
networks \cite{CalvoZaragoza2016d}.
Anecdotal evidence from ongoing experiments suggest that there is little
difference between handwritten and printed notation in this respect,
so the problem of uneven document lighting should be solvable
separately, with synthesized notation and equally synthesized lighting
for training data, and therefore is not an important concern in designing
our dataset.

Page deformation, including 3-D displacement, can be synthesized as well,
by mapping the flat surface on a generated ``landscape'' and simulating
perspective. Therefore, again, our dataset does not necessarily need to include
images of scores with deformations.

A separate concern for historical documents is also background paper degradation
and bleedthrough. While paper textures can be successfully simulated,
realistic models of bleedthrough are, to the best of our knowledge, not available.

Also bearing in mind that the dataset is supposed to address stages 2 and 3
of the OMR pipeline, and the extent to which difficulties primarily tackled 
at the first stage of the OMR pipeline (image processing) can be simulated,
leads us to believe that {\bf it is a reasonable choice to annotate binary images.}

\subsection{Topological consistency}
\label{subsec:choiceofdata:topologicalconsistency}

The tightness of spacing in \cite{Byrd2015} refers to default
horizontal and vertical distances between musical symbols. In printed music,
this is a parameter of typesetting that may also change dynamically in order
to place line and page breaks for musician comfort. The problem with tight
spacing is that symbols start encroaching into each others' bounding boxes,
and horizontal distances change while vertical distances don't,
thus perhaps breaking assumptions about relative notation spacing.
Byrd and Simonsen
give an example where the augmentation dot of a preceding note is in a position
where it can be easily confused with a staccato dot of its following note
(\cite{Byrd2015}, fig. 21).
This leads to increasingly ambiguous inputs to the primitive assembly stage.

However, in handwritten music, variability in spacing is superseded by the variability
of handwriting itself, which itself introduces severe problems with respect
to spatial relationships of symbols. 
Most importantly, handwritten music
gives no topological constraints: by definition straight lines, 
such as stems, become curved,
noteheads and stems do not touch, accidentals and noteheads {\em do} touch,
etc. However, some writers are more disciplined than others in this respect.
The various styles of handwriting, and the ensuing challenges, also have to be
represented in the dataset, as broadly as possible. As there is currently no
model available to synthesize handwritten music scores, we need to cover this
spectrum directly through choice of data.

We find {\bf adherence to topological standards}
to be a more general term that describes this particular class of difficulties.
Tightness of spacing is a factor that globally influences topological standardization
in printed music; in handwritten music, the variability in handwriting styles
is the primary source of inconsistencies with respect to the rules 
of CWMN topology. This variability includes the difference
between contemporary and historical music handwriting styles.

To summarize, it is essential to include the following challenges into the dataset
directly through choice of the musical score images:

\begin{itemize}[noitemsep]
\item Notation complexity,
\item Handwriting styles,
\item Realistic historical document degradation.
\end{itemize}

\section{Existing datasets}
\label{sec:existingdatasets}

There are already some OMR datasets for handwritten music.
What tasks are they for? Can we save ourselves some work
by building on top of them?
Is there a dataset that perhaps already satisfies 
the requirements of sections \ref{sec:groundtruth} and \ref{sec:choiceofdata}?

Reviewing subsec. \ref{subsec:groundtruth:stages},
the subtasks at stages 2 and 3 of the OMR pipeline are:

\begin{itemize}[noitemsep]
\item Staffline removal
\item Symbol localization
\item Symbol classification
\item Symbol assembly
\end{itemize}

\noindent
(See table \ref{tab:groundtruth:io} for input and output definitions.)
How are these tasks served by datasets of handwritten music scores?

\subsection{Staff removal}
\label{subsec:existingdatasets:staffremoval}

For staff removal in handwritten
music, the best-known and largest dataset is CVC-MUSCIMA \cite{Fornes2012},
consisting of 1000 handwritten scores (20 pages of music, 
each copied by hand by 50 musicians). The dataset is distributed 
with further 11 pre-computed distortions for each of these scores
according to Dalitz et al. \cite{Dalitz2008} for a total of 12000 images.
The state-of-the-art for staff removal has been 
established with a competition using CVC-MUSCIMA.
\cite{fornes2011icdar}

For each input image, CVC-MUSCIMA has three binary images:
a ``full image''
mask,\footnote{Throughout this work, we use the term
{\em mask} of some graphical element $s$ to denote a binary matrix 
that is applied by elementwise
multiplication (Hadamard product): a value of $0$ means
``this pixel does not belong to $s$'', a value of $1$ means ``this pixel
belongs to $s$''. A mask of all non-staff symbols therefore
has a $1$ for each pixel such that it is in the foreground,
and belongs to one or more symbols that are not stafflines, 
and a $0$ in all other cells, and if opened in an image viewer,
it looks like white-on-black results of staffline removal.}
which contains all foreground pixels,
a ``ground truth'' mask of all pixels that belong to a staffline and at
the same time to no other symbol, and a ``symbols'' mask
that complementarily contains only pixels that belong to some {\em other}
symbol than a staffline. The dataset was collected by giving a set
of 20 pages of sheet music to 50 musicians. 
Each was asked to rewrite the same 20 pages by hand,
using their natural handwriting style. A standardized paper and pen 
was provided for all the writers, so that binarization and staff removal
was done automatically with very high accuracy, and the results
were manually checked and corrected. 
In vocal pieces, lyrics were not transcribed.

Then, the 1000 images were run through the 11 distortions defined
by Dalitz et al. \cite{Dalitz2008}, to produce a final set of 12 000
images that are since then used to evaluate staff removal algorithms,
including a competition at ICDAR \cite{fornes2011icdar}.

\begin{figure}[!t]
\centering
\subfloat[Writer 9: nice handwriting.]{\includegraphics[width=3.0in]{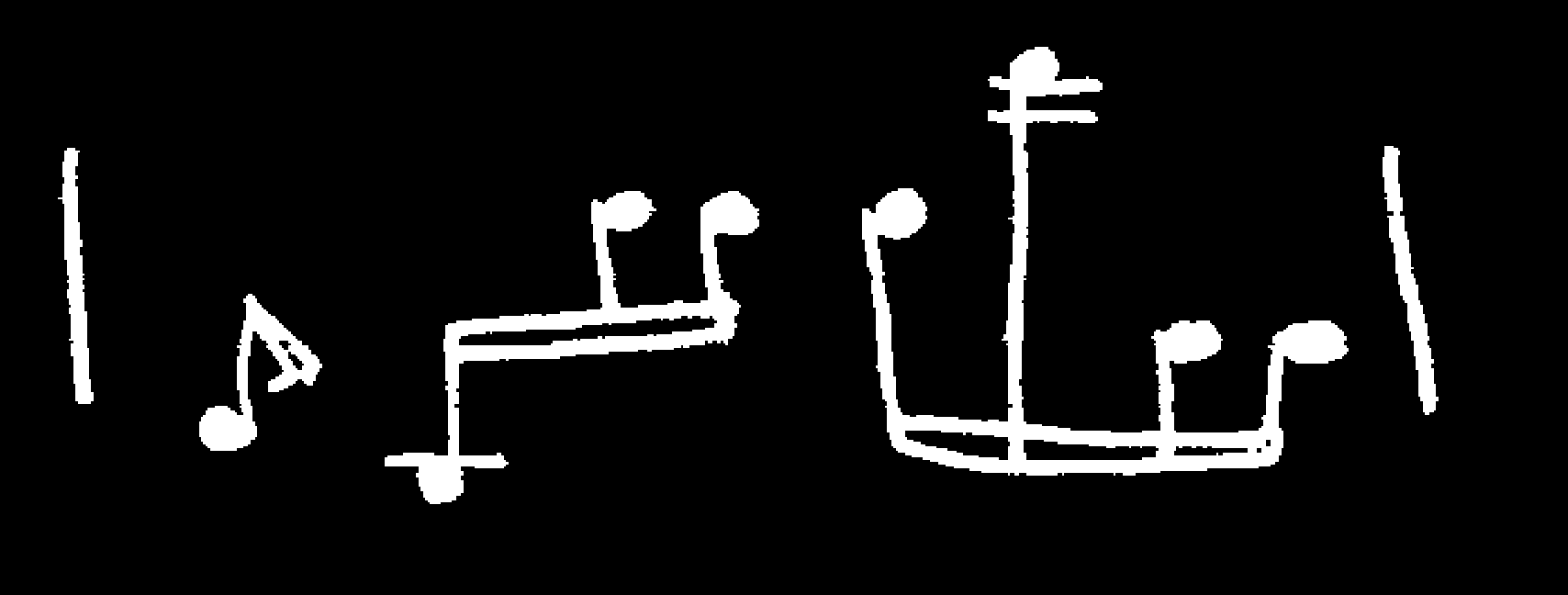}%
\label{fig:handwriting:nice}}
\vfil
\subfloat[Writer 49: Disjoint notation primitives]{\includegraphics[width=3.0in]{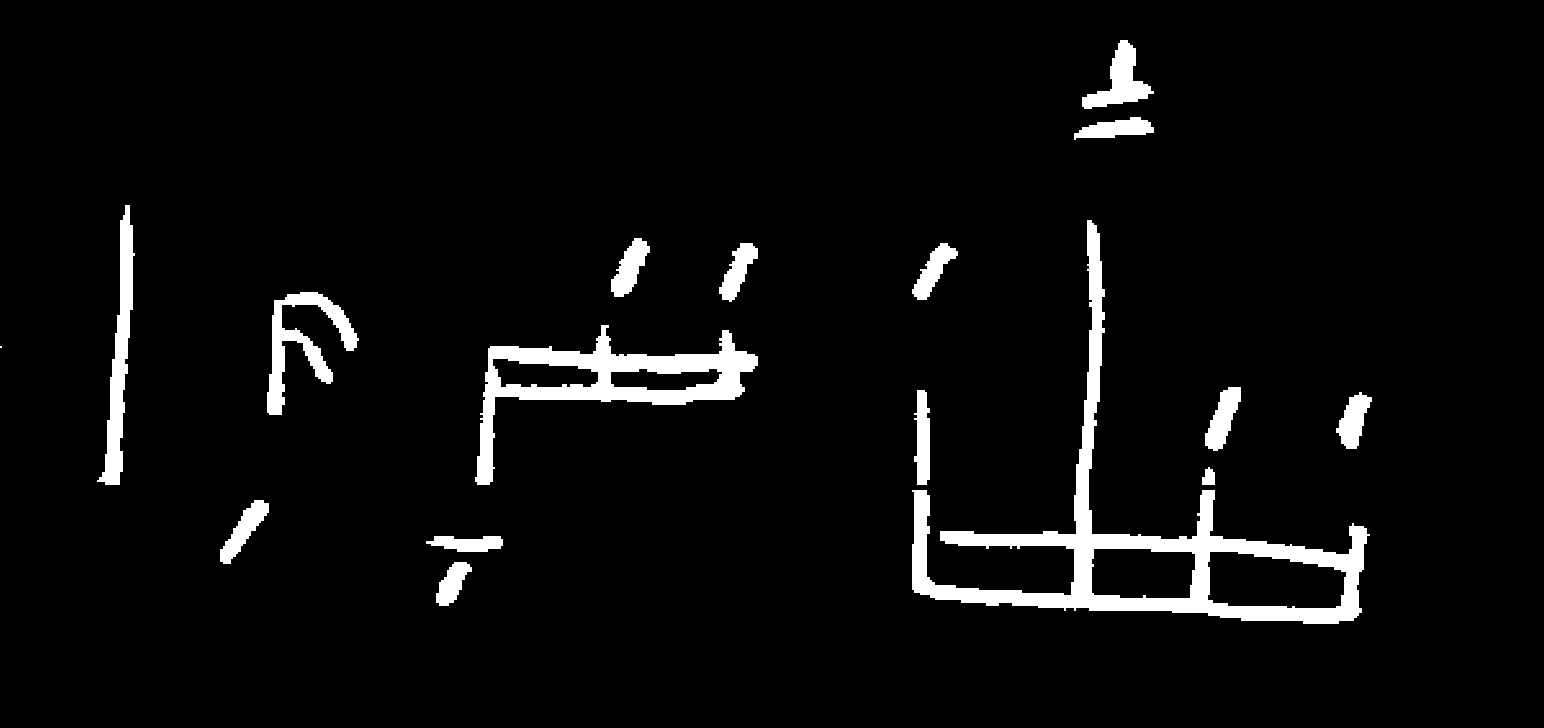}%
\label{fig:handwriting:ugly}}
\vfil
\subfloat[Writer 22: Disjoint primitives and deformed noteheads. Some noteheads will be very hard to distinguish from the stem]{\includegraphics[width=3.0in]{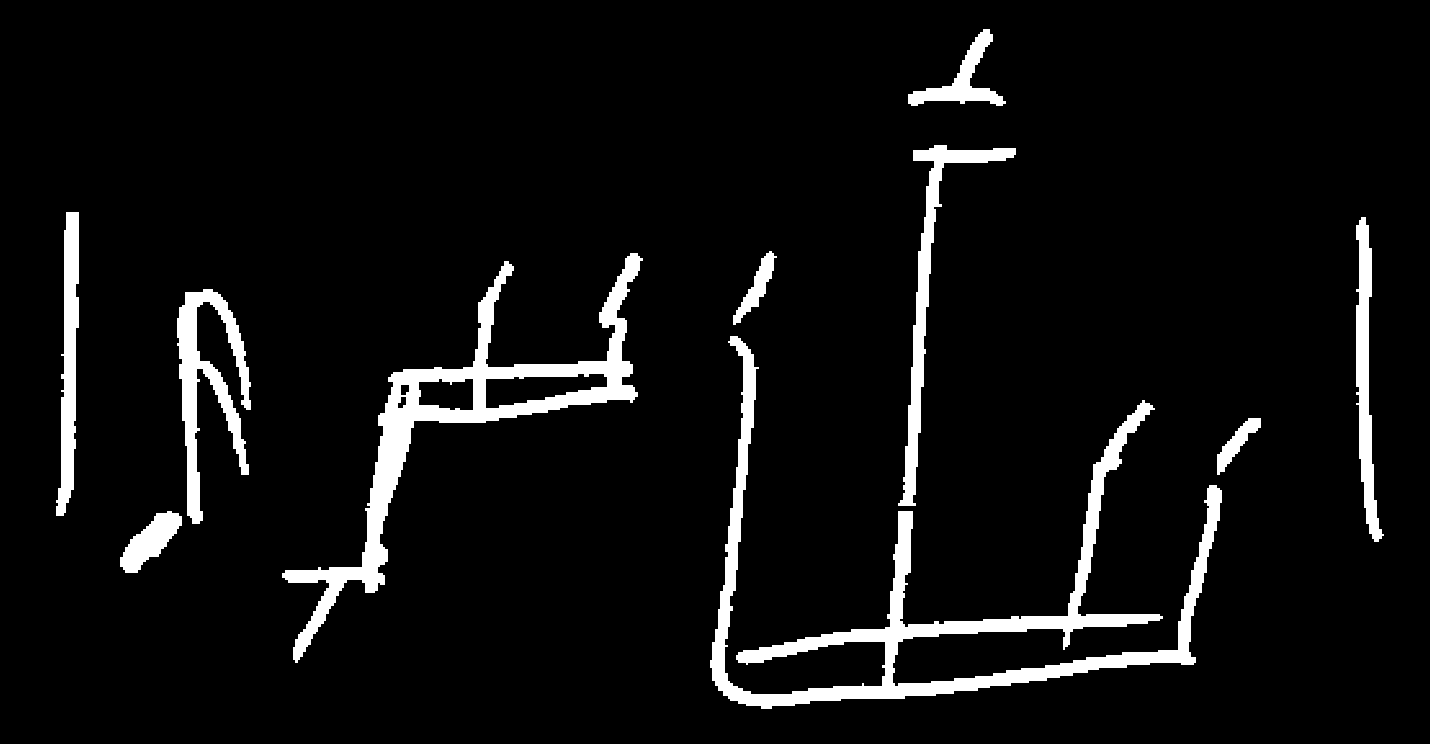}
\label{fig:handwriting:reallyugly}}
\caption{Variety of handwriting styles iv the CVC-MUSCIMA dataset.}
\label{fig:handwriting}
\end{figure}

Which of our requirements does the dataset fulfill? With respect to ground truth,
it provides staff removal, which has proven to be quite time-intensive to annotate
manually, but no symbol annotations (it has been mentioned
in \cite{Fornes2012} as future work). The dataset also fulfills the requirements
for a good choice of data, as described in \ref{sec:choiceofdata}. With respect
to notation complexity, the 20 pages of CVC-MUSCIMA include
scores of all 4 levels, as summarized in table \ref{tab:existingdatasets:cvcmuscima}
(some scores are very nearly single-voice per staff; these are marked with a (?)
sign to indicate their higher category is mostly undeserved).
There is also a wide array of music notation symbols across the 20 pages,
including tremolos, glissandi, an abundance of grace notes, ornaments, trills;
clef changes, time and key signature changes, and even voltas and less common
tuples are found among the music. Handwriting style varies greatly
among the 50 writers, including topological inconsistencies: some writers
write in a way that is hardly distinguishable from printed music, some
write rather extremely disjoint notation and short diagonal lines instead of
round noteheads, as illustrated in fig. \ref{fig:handwriting}.

The images are provided as binary, so there is no variability on the
 image quality axis of difficulty, However, in \ref{subsec:choiceofdata:imagequality},
 we have determined this to be acceptable. Another limitation
 is that all the handwriting is contemporary.
Finally, and importantly, it is freely available for download 
under a Creative Commons NonCommercial Share-Alike 4.0
license.\footnote{\url{http://www.cvc.uab.es/cvcmuscima/index_database.html}}

\begin{table}[!t]
\renewcommand{\arraystretch}{1.3}
\caption{CVC-MUSCIMA notation complexity}
\label{tab:existingdatasets:cvcmuscima}
\centering
\begin{tabular}{c|c}
{\bf Complexity level} & {\bf CVC-MUSCIMA pages} \\
\hline
\hline
Single-staff, single-voice & 1, 2, 4, 6, 13, 15, \\
Single-staff, multi-voice & 7, 9 (?), 11 (?), 19 \\
Multi-staff, single-voice & 3, 5, 12, 16, \\
Multi-staff, multi-voice & 14 (?), 17, 18, 20 \\
Pianoform & 8, 10 \\
\end{tabular}
\end{table}

\subsection{Symbol classification and localization}
\label{subsec:existingdatasets:classificationandlocalization}

Handwritten symbol classification systems can be trained
and evaluated on the HOMUS dataset of Calvo-Zaragoza and Oncina
\cite{CalvoZaragoza2014},
which provides 15200 handwritten musical symbols (100 writers,
32 symbol classes, and 4 versions of a symbol
per writer per class, with 8 for note-type symbols).
HOMUS is also interesting in that the data is captured
from a touchscreen: it is availabe in {\em online} form,
with $x, y$ coordinates for the pen at each time slice of $16$ ms,
and for offline recognition (ie. from a scan), images of music symbols
can be generated from the pen trajectory. Together with
potential data from a recent multimodal recognition (offline + online)
experiment \cite{CalvoZaragoza2016c}, these datasets might
enable trajectory reconstruction from offline inputs. Since online
recognition has been shown to perform better than offline on the dataset
\cite{CalvoZaragoza2014}, such a component -- if performing well --
could lead to better OMR accuracy.

Other databases of isolated musical symbols have been collected, 
but they have not been made available. 
Silva \cite{Silva2013Thesis} collects a dataset from 50 writers
with 84 different symbols, each drawn 3 times, for a total of 12600 symbols.

However, these datasets only contains isolated symbols,
not their positions on a page. While it might be possible to synthesize
handwritten music pages from the HOMUS symbols,
such a synthetic dataset will be rather limited, as HOMUS does not contain
beamed groups and chords.\footnote{It should also be noted that HOMUS
is not available under an open license, so copyright restrictions apply.
Not even every EU country has the appropriate ``fair use''-like exception for
academic research, even though it is mandated by the EU Directive 2001/29/EC,
Article 5(3) (\url{http://eur-lex.europa.eu/legal-content/EN/ALL/?uri=CELEX:32001L0029}).}

For symbol localization (as well as classification),
we are only aware of a dataset of 3222 handwritten symbols
by the group of Rebelo et al.
\cite{Rebelo2010, Rebelo2012Thesis}. 
This dataset is furthermore only available upon request,
not publicly.

\subsection{Notation reconstruction and final representation}
\label{subsec:existingdatasets:reconstruction}

We are not aware of a dataset that explicitly marks
the relationships among handwritten musical symbols.

Musical content reconstruction is usually understood to entail
deriving pitch and relative duration of the written notes.
It is possible to mine early music manuscripts in the IMSLP
database\footnote{\url{https://www.imslp.org}} and pair them
against their open-source editions, which are sometimes provided on the
website as well, or look for matching encoded data in large repostories 
such as Mutopia\footnote{\url{http://www.mutopiaproject.org}} 
or KernScores\footnote{\url{http://humdrum.ccarh.org}}; however,
we are not aware of such a paired collection for OMR, or any other
available dataset for pitch and duration reconstruction.
While Bellini et al. \cite{Bellini2007} do perform evaluation of OMR systems
with respect to pitch and duration on a limited dataset of 7
pages, the evaluation was done manually, without creating a ground truth
for this data.

\section{The MUSCIMA++ dataset}
\label{sec:data}

We finally describe the data that MUSCIMA++~\currentversion~makes available.

Our main source of musical score images is the CVC-MUSCIMA dataset
described in subsection \ref{subsec:existingdatasets:staffremoval}.
The goal for the first round of MUSCIMA++ annotation was for each 
of our annotators to mark one of the 50 versions for each of the 20 pages.
With 7 available annotators, this amounted to
140 annotated pages of music. Furthermore, we assigned the 140
out of 1000 pages of CVC-MUSCIMA so that all of the 50 writers
are represented as equally as possible: 2 or 3 pages are annotated
from each writer.\footnote{With the exception of writer 49, who
is represented in 4 images due to a mistake in the distribution workflow 
that was only discovered after the image was already annotated.}

There is a total of~\numberofsymbols~symbols
marked in the 140 annotated pages of music, of 107 distinct symbol classes.
There are~\numberofrelationships~relationships between pairs of symbols.
The total number of {\em notes} encoded in the dataset is~\numberofnotes.
The set of symbol classes consists of both 
notation primitives, such as noteheads or beams, and higher-level notation objects, 
such as key signatures or time signatures. The specific choices of symbols
and ground truth policies is described
in subsec. \ref{subsec:data:groundtruth}.

The frequencies of the most significant symbols are described
in table \ref{tab:data:symbolfreqs}. We can draw two lessons immediately
from the table. First, even when lyrics are stripped away \cite{Fornes2012},
texts make up a significant portion of music
notation -- nearly 5 \% of symbols are letters. Some utilization of handwritten
OCR, or at least identifying and removing texts, therefore seems
reasonably necessary.
Second, at least among 18th to 20th
century music, some 90 \% of notes occur as part of a beamed group,
so works that do not tackle beamed groups are in general greatly restricted
(possibly with the exception of choral music such as hymnals, where 
isolated notes are still the standard).

How does MUSCIMA++ compare to existing datasets?
Given that individual notes are split into primitives and other
ground truth policies, to obtain a fair comparison,
we should subtract the stem count, letters and texts, and
the {\tt measure\_separator} symbols. Some sharps and flats
are also part of key signatures, and numerals are part of time signatures,
which again leads to two symbols where other datasets may only annotate
one. These subtractions bring us to a more directly comparable symbol count
of about 57000.

\begin{table}[!t]
\renewcommand{\arraystretch}{1.3}
\caption{Symbol frequencies in MUSCIMA++}
\label{tab:data:symbolfreqs}
\centering
\begin{tabular}{lr|lr}
{\bf Symbol} & {\bf Count} & {\bf Symbol (cont.)} & {\bf Count} \\
\hline
\hline
stem & 21416 & 16th\_flag & 495 \\
notehead-full & 21356 & 16th\_rest & 436 \\
ledger\_line & 6847 & g-clef & 401 \\
beam & 6587 & grace-notehead-full & 348 \\
thin\_barline & 3332 & f-clef & 285 \\
measure\_separator & 2854 & other\_text & 271 \\
slur & 2601 & hairpin-decr. & 268 \\
8th\_flag & 2198 & repeat-dot & 263 \\
duration-dot & 2074 & tuple & 244 \\
sharp & 2071 & hairpin-cresc. & 233 \\
notehead-empty & 1648 & half\_rest & 216 \\
staccato-dot & 1388 & accent & 201 \\
8th\_rest & 1134 & other-dot & 197 \\
flat & 1112 & time\_signature & 192 \\
natural & 1089 & staff\_grouping & 191 \\
quarter\_rest & 804 & c-clef & 190 \\
tie & 704 & trill & 179 \\
key\_signature & 695 & {\em All letters} & {\em 4072} \\
dynamics\_text & 681 & {\em All numerals} & {\em 594} \\

\end{tabular}
\end{table}

{\em A note on naming conventions:} CVC-MUSCIMA refers to the set
of binary images described in \ref{subsec:existingdatasets:staffremoval}.
MUSCIMA++~\currentversion~refers to the symbol-level ground truth
of the selected 140 pages. (Future versions of MUSCIMA++ may contain
more types of data, such as MEI-encoded musical information, or other
representations of the encoded musical semantics.)
The term ``MUSCIMA++ images'' refers to those 140 undistorted images
from CVC-MUSCIMA that have been annotated so far.

\subsection{Designated test sets}
\label{subsec:data:testsets}

To give some basic guidelines on comparing trainable systems 
over the dataset, we designate some images to serve as a test set.
One can always use a different train-test split; however, we believe our
choice is balanced well. Similar to \cite{CalvoZaragoza2014}, we provide 
a {\em user-independent} test set, and a {\em user-dependent} one.
Each of these contains 20 images, one for each CVC-MUSCIMA page.
However, they differ in how each of these 20 is chosen from the 7
available versions, with respect to the individual writers.

The {\bf user-independent} test set evaluates how the system handles 
data form previously unseen writers.
The images are split so that the 2 or 3
MUSCIMA++ images from any particular writer are either all in the training
portion, or all in the test portion of the data.

The {\bf user-dependent} test set, to the contrary, contains at most
one image from each writer in its set of the 20 CVC-MUSCIMA pages.
For each writer in the user-dependent test set, there is also at least one
image in the training data. This allows experimenting with at least some
amount of user adaptation.

Furthermore, both test sets are chosen so that the annotators
are represented as uniformly as possible, so that the evaluation is not
biased towards the idiosyncracies of a particular annotator.\footnote{The
choice of test set images is provided as supplementary data, together with
the dataset itself.}

\subsection{MUSCIMA++ ground truth}
\label{subsec:data:groundtruth}

How does MUSCIMA++ implement the requirements described in \ref{sec:groundtruth}?

Our ground truth is a graph.
We define a fine-grained vocabulary of musical symbols as its vertices,
and we define relationships between symbols to be expressed as 
unlabeled directed edges.
We then define how ordered pairs participate in relationships
(noteheads connect to accidentals, numerals combine to form time signatures,
etc.)\footnote{The most complete
annotation guidelines detailing what the symbol set is and how to deal with
individual notations are available online: 
\url{http://muscimarker.readthedocs.io/en/develop/instructions.html}}

Each vertex (symbol) furthermore has a set of attributes. These are
a superset of the primitive attributes in \cite{Miyao2000}.
For each symbol, we encode:

\begin{itemize}[noitemsep]
\item its {\bf label} ({\tt notehead}, {\tt sharp}, {\tt g-clef}, etc.),
\item its {\bf bounding box} with respect to the page,
\item its {\bf mask:} exactly which pixels in the bounding box belong to this symbol?
\end{itemize}

\noindent
The mask is especially important for beams, 
as they are often slanted and so their bounding box overlaps with
other symbols (esp. stems and parallel beams). Slurs also often have this problem.
Annotating the mask enables us to build an accurate model of actual symbol shapes.

The symbol set includes what \cite{Droettboom2004}, 
\cite{Bellini2007} and \cite{Byrd2015}
would describe as a mix of low-level symbols as well as high-level symbols,
but without explicitly labeling the symbols as either. Instead of trying to
categorize symbols according to whether they carry semantics or not,
we chose to express the
high- vs. low-level dichotomy through the rules for forming relationships.
This leads to ``layered'' annotation. 
For instance, a 3/4 time signature is annotated
using three symbols: a {\tt numeral\_3}, {\tt numeral\_4}, and a {\tt time\_signature}
symbol that has outgoing relationships to both of the numerals involved.
In MUSCIMA++~\currentversion~we
do {\em not} annnotate invisible symbols (e.g. implicit tuplets). 
Each symbol has to have at least one foreground pixel.

\begin{figure}[!t]
\centering
\includegraphics[width=3.0in]{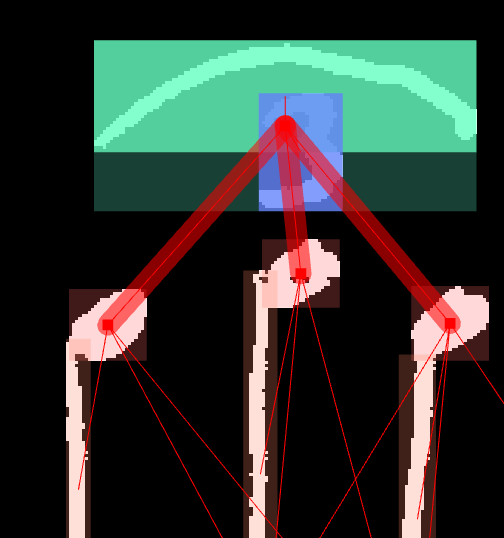}%
\caption{Two-layer annotation of a triplet. Highlighted symbols are {\tt numeral\_3},
 {\tt tuple\_bracket/line}, and the three noteheads that form the triplet. The
 {\tt tuple} symbol itself, to which the noteheads are connected, is the darker
 rectangle encompassing its two components; it has relationships leading to both
 of them (not highlighted).}
\label{fig:data:multilayertuple}
\end{figure}

The policy for making decisions that are arbitrary with respect to the information
content, as discussed at the end of subsec. \ref{subsec:groundtruth:requirements},
was set to stay as close as possible to the {\em written page}, rather than
the semantics. If this principle was in conflict with the requirement
for both reprintability and replayability introduced in \ref{sec:introduction},
 a symbol class was added to the vocabulary to capture the requisite meaning.
 Examples are the {\tt key\_signature}, {\tt time\_signature}, {\tt tuple},
 or {\tt measure\_separator}. These second-layer symbols are often composite,
 but not necessarily so: for instance, a single {\tt sharp} can also form
 a {\tt key\_signature}, or a {\tt measure\_separator} is expressed by
 a single {\tt thin\_barline}.  An example of this structure for
 is given in figure \ref{fig:data:multilayertuple}.
 
While we take care to define relationships so that the result is a Directed
Acyclic Graph (DAG), there is no hard limit on the maximum oriented path length.
However, in practice, it very rarely goes deeper than 3, 
as the path in the {\tt tuple} example
leading from a {\tt notehead} to the {\tt tuple} and then to its constitutent
{\tt numeral\_3} or {\tt tuple\_bracket/line}, and in most cases, this depth
is at most 2.
 
On the other hand, we do not break down symbols that consist of multiple
connected components, unless it is possible that these components
can be seen in valid music notation in various configurations.
An empty notehead may show up with a stem, without one, with multiple
stems when two voices share pitch,\footnote{As seen in page 20 of CVC-MUSCIMA.}
or it may share stem with others, so we define these as separate symbols. 
An f-clef dot cannot exist without the rest of the clef, 
and vice versa, so we define the {\tt f-clef} as a single symbol;
on the other hand, a single repeat spanning multiple staves may have
a variable number of repeat dots, so we define a {\tt repeat-dot} separately.
This is a different policy from
Miyao \& Haralick \cite{Miyao2000}, who split e.g. the {\tt repeat\_measure}
``percent sign'' into three primitives.

{\bf We do not define a {\tt note} symbol.}
Notes are hard to pin down on the written page: in the traditional understanding
of what is a "note" symbol \cite{Rebelo2012Thesis} \cite{CalvoZaragoza2014}
\cite{Droettboom2004},
they consist of multiple primitives (notehead and stem and beams or flags),
but at the same time, multiple notes can {\em share} these primitives,
including noteheads --
the relationship between high- and low-level symbols has in general an {\tt m:n}
cardinality. Another high-level symbol may be a key signature, which can consist
of multiple sharps or flats. It is not clear how to annotate notes. 
If we follow the "semantics" criterion for distinguishing between the low-
and high-level symbol description of the written page, 
should e.g. an accidental be considered a part of the note, 
because it directly influences its pitch?\footnote{
We would go as far as to say that it is inadequate
to try marking "note" graphical objects
in the musical score. A note is a basic unit of {\em music}, but it is not a unit
of music {\em notation}. Music notation {\em encodes} notes,
it does not {\em contain} them.}

The policy for defining how relationships should be formed
was to make noteheads independent of each other.
That is: as much as possible of the semantics of a note corresponding
to a notehead can be inferred based on the explicit relationships 
that pertain to this particular notehead.
However, this ideal is not fully implemented
in MUSCIMA++~\currentversion, and possibly
cannot be reasonably reached, given the rules of music notation.
While it is possible to infer duration from the current annotation
(with the exception of implicit tuples), not so much pitch. First of
all, one would need to add staffline and staff objects and
link the noteheads to the staff. This is not explicitly done in
MUSCIMA++~\currentversion, but given the staff removal ground
truth included with the annotated CVC-MUSCIMA images, it should be
merely a technical step that does not require much manual annotation.
The other missing piece of the pitch puzzle are assignment of notes 
to measures, and precedence relationships. Precedence relationships
need to include (a) notes, to capture effects of accidentals
at the measure scope; (b) clefs and key signatures, which can change
within one measure,\footnote{Even though this does not occur 
in CVC-MUSCIMA, it does happen, as illustrated e.g. by Fig. 8 in \cite{Byrd2015}.}
so it is not sufficient to attach them to measures.

Finally, in cases where these policy considerations did not provide
clear guidance, we made choices in the ground truth definition
based on aesthetics of the annotation interface, so that annotators
could work faster and more accurately.

\subsection{Available tools}
\label{subsec:data:tools}

In order to make using the dataset easier, we provide two software tools
under an open license.

First, the {\tt musicma} Python 3 
package\footnote{\url{https://github.com/hajicj/muscima}}
implements the MUSCIMA++ data
model, which can parse the dataset and enables manipulating the data
further (such as assembling the related primitives into notes, to provide 
a comparison to the existing datasets with different symbol sets).

Second, we provide the {\tt MUSCIMarker} application.\footnote{\url{https://github.com/hajicj/MUSCIMarker}} This is the annotation interface
used for creating the dataset, and it can visualize the data.

\subsection{Annotation process}
\label{subsec:data:annotation}

\begin{figure}[!t]
\centering
\includegraphics[width=3.3in]{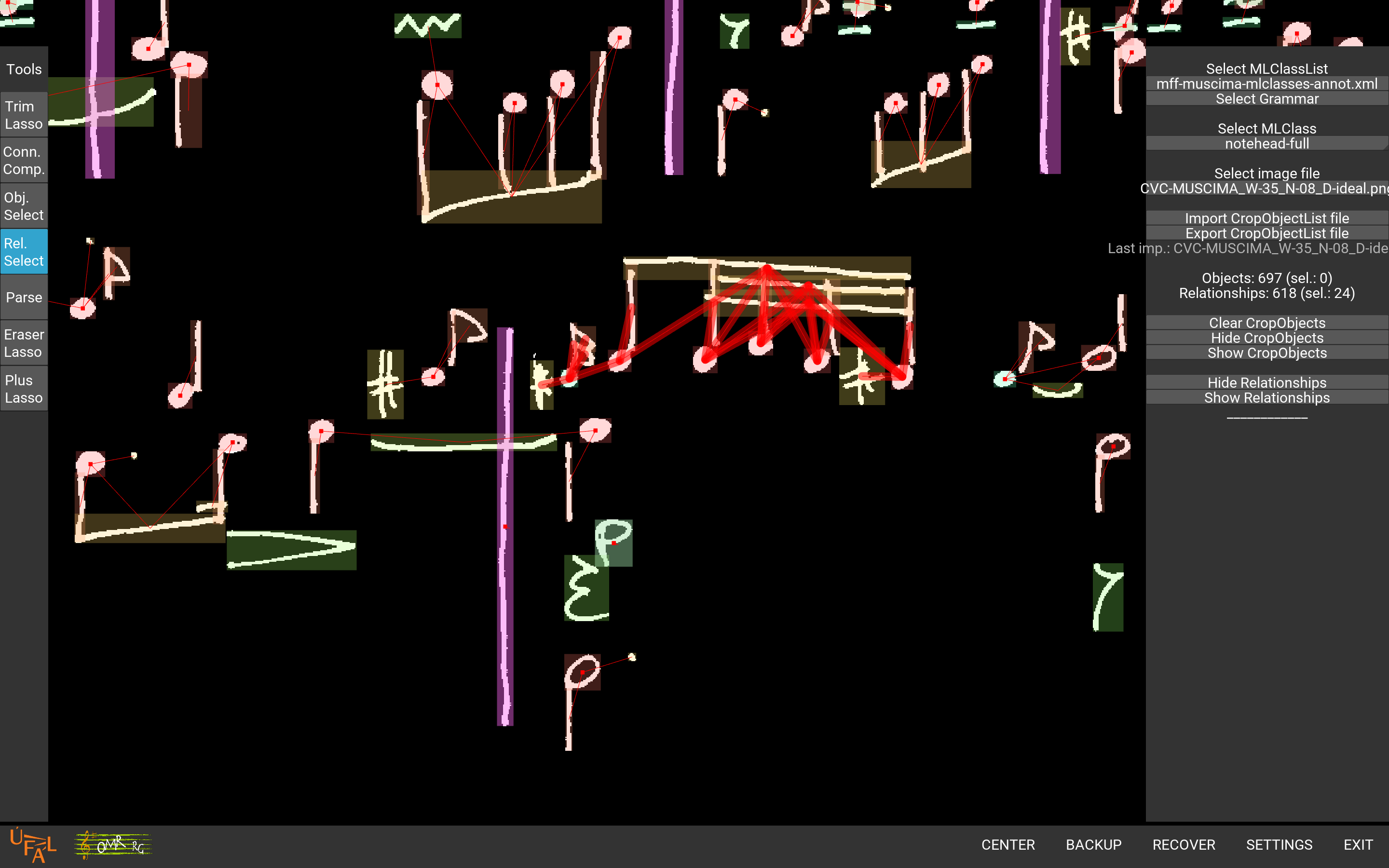}%
\caption{MUSCIMarker 1.1 interface. Tool selection on the left; file controls on the right. Highlighted relationships have been selected. (Last bar of from MUSCIMA++
image {\tt W-35\_N-08}.)}
\hfil
\label{fig:data:muscimarker}
\end{figure}

Annotations were done using custom-made
MUSCIMarker open-source software, version 1.1.\footnote{\url{https://lindat.mff.cuni.cz/repository/xmlui/handle/11234/1-1850}, ongoing development 
at \url{https://github.com/hajicj/MUSCIMarker}}
The annotators worked on symbols-only CVC-MUSCIMA images,
which allowed for more efficient annotation.
The interface used to add symbols consisted of two tools:
a background-ignoring lasso selection, and
connected component selection.\footnote{Basic usage is described in the MUSCIMarker tutorial: \url{http://muscimarker.readthedocs.io/en/develop/tutorial.html}} A screenshot of the annotation interface is 
in fig. \ref{fig:data:muscimarker}.

As annotation was under way, due to
the prevalence of first-try inaccuracies, we added editing tools
that enabled annotators to "fine-tune" the symbol shape by adding
or removing arbitrary foreground regions to the symbol's mask.
Adding symbol relationships was done by another lasso selection
tool. An important speedup was also achieved by providing a rich set of
keyboard shortcuts. The effect was most acutely felt 
in the keyboard interface for assigning labels to symbols,
as the vocabulary of available labels is rather extensive (\numberofclasses,
of which \numberofclassesused~are actually present in the data).

There were seven annotators working on MUSCIMA++. Four of these are
professional musicians, three are experienced amateur musicians.
The annotators were asked to complete three progressively more complex
training examples. There were no IT skills required (we walked them through
MUSCIMarker installation and usage, the odd troubleshooting was done through
the TeamViewer\footnote{\url{https://www.teamviewer.com}} remote interface).
Two of the training examples were single measures, for basic familiarization
with the user interface; the third was a full page, to ensure understanding
of the majority of notation situations. Based on this "training round",
we have also further refined the ground truth definitions.

As noted above, each annotator completed one image 
of each of the 20 CVC-MUSCIMA pages.
The work was dispatched to annotators
in four packages of 5 images each, one package at a time.
After each package submission by an annotator, we checked for correctness.
Automated validation was implemented in MUSCIMarker
to check for ``impossible'' or missing
relationships (e.g.: a stem and a beam should not be connected; 
a grace notehead has to be connected to a ``normal'' notehead).
However, there was still need for manual checks and manually correcting
mistakes found in auto-validation, as the validation itself was an advisory
voice to highlight questionably annotated symbols, not an authoritative response.

Manually checking each submitted
file also allowed for continuous annotator training and filling in
``blind spots'' in the annotation guidelines (such as specifying how to deal
with {\em volta} signs, or tuples). 
Some notation events were simply not anticipated
(e.g., mistakes that CVC-MUSCIMA writers made in transcription or non-standard
symbols).
Feedback was provided individually after each package was submitted 
and checked. This feedback was the main mechanism for continuous training.
Requests for clarifications of guidelines for situations that proved
problematic were also disseminated to the whole group.

At first, the quality of annotations was inconsistent: some annotators
performed well, some poorly. Some required extensive feedback.
Thanks to the continuous communication
and training, the annotators improved, and the third and fourth packages required
relatively minor corrections. Overall, however, only one annotator submitted
work at a quality that required practically no further changes during quality control.
Differences in annotator speed did not equalize as much as annotation correctness.

Finally, after collecting annotations for all 140 images, we performed
a second quality control round, this time with further automated checks.
We checked for disconnected symbols, and for symbols with suspiciously sparse
masks (a symbol was deemed suspicious if more than $0.07$ 
of the foreground pixels
in its bounding box were not marked as part of any symbol at all).
This second round of quality control uncovered yet more inaccuracies,
and we also fixed other clearly wrong markings (e.g., if a significant
amount of stem-only pixels was marked as part of a beam).

On average throughout the annotations, the fastest annotator managed to mark about $6$ symbols
per minute, or one per $10$ seconds; the next fastest came at $4.6$ symbols
per minute. Two slowest annotators clocked in around $3.4$ symbols per minute.
The average speed overall was $4.3$ symbols per minute, or one
per $14$ seconds. The ``upper bound'' on annotation speed was established
by the first author (who is intimately familiar with the most efficient ways of using
the MUSCIMarker annotation tool) to be $8.9$ objects per minute (one in $6.75$
seconds). These numbers are computed over the whole time spent annotating,
so they include the periods during which annotators were marking relationships
and checking their work: in other words, if you plan to extend the dataset
with a comparable annotator workforce, you can expect
an average page of about 650 symbols to take about $2 \frac{3}{4}$ hours.

Annotating the dataset using the process detailed above took roughly
400 hours of work; the ``quality control'' correctness checks took
an additional 100 -- 150. The second, more complete round of quality
control took roughly 60 -- 80 hours, or some 0.5 hours 
per image.\footnote{For the sake of completeness,
implementing MUSCIMarker took about 600 hours,
including the learning curve for the GUI framework.}

\subsection{Inter-annotator agreement}
\label{subsec:data:agreement}

In order to assess (a) whether the annotation guidelines are well-defined,
and (b) the extent to which we can trust annotators, we conducted a test:
all seven annotators were given the same image to annotate, and 
we measured inter-annotator agreement.
Inter-annotator agreement does explicitly not decouple the factors (a) and (b).
However, given that the overall expected level of ambiguity is relatively
low, and given the learning curve along which the annotators were moving
throughout their work, which would as be hard to decouple from genuine
(a)-type disagreement, we opted to not expend resources on annotators
re-annotating something which they had already done, and therefore
cannot provide exact intra-annotator agreement data.

Another use of inter-annotator agreement is to provide an upper bound
on system performance. If a system performs better than average inter-annotator
agreement, it may be overfitting the validation set. (On the other hand, it may have
merely learned to compensate for annotator mistakes -- more analysis is needed
before concluding that the system overfits. But it is a useful warning: one
should investigate unexpectedly high performance numbers.)

\subsubsection{Computing agreement}

In order to evaluate the extent to which two annotators agree on how
a given image should be annotated, we perform two steps:

\begin{itemize}[noitemsep]
\item Align the annotated object sets against each other,
\item Compute the macro-averaged f-score over the aligned object pairs.
\end{itemize}

\noindent
Objects that have no counterpart contribute $0$ to both precision
and recall.

Alignment was done in a greedy fashion. For symbol sets $S, T$,
we first align each $t \in T$ to the $s \in S$ with the highest
pairwise f-score $F(s, t)$, then vice versa align each $s \in S$ to the
$t \in T$ with the highest pairwise f-score. Taking the intersection,
we then get symbol pairs $s, t$ such that they are each other's
``best friends'' in terms of f-score. The symbols that do not have such
a clear counterpart are left out of the alignment. Furthermore, symbol
pairs that are not labeled with the same symbol class are removed
from the alignment as well.

When breaking ties in the pairwise matchings (from both directions),
symbol classes $c(s), c(t)$ are used. If $F(s, t_1) = F(s, t_2)$,
but $c(s) = c(t_1)$ while $c(s) \neq c(t_2)$,
then $(s, t_1)$ is taken as an alignment candidate instead
of $(s, t_2)$. (If both $t_1$ and $t_2$ have the
same class as $s$,
then then the tie is broken randomly. In practice, this would be extremely
rare and would not influence agreement scores very much.)

\subsubsection{Agreement results}

The resulting f-scores are summarized in table \ref{tab:data:agreement}.
We measured inter-annotator agreement both before and after quality
control (noQC-noQC and withQC-withQC),
and we also measured the extent to which quality control changed
the originally submitted annotations (noQC-withQC).
Tentatively, the post-QC measurements
reflect the level of genuine disagreement among the annotators about
how to lead the boundaries of objects in intersections and the inconsistency
of QC, while the pre-QC measurements also measures the extent of
actual mistakes that were fixed in QC.

\begin{table}[!t]
\renewcommand{\arraystretch}{1.3}
\caption{Inter-annotator agreement}
\label{tab:data:agreement}
\centering
\begin{tabular}{c|cc}
Setting & macro-avg. f-score \\
\hline
\hline
noQC-noQC (inter-annot.) & 0.89 \\
noQC-withQC (self) & 0.93 \\
withQC-withQC (inter-annot.) & 0.97 \\
\end{tabular}
\end{table}

Ideally, the task of annotating music notation symbols is relatively unambiguous.
Legitimate sources of disagreement lie in two factors:
unclear symbol boundaries in intersections, and illegible handwriting.
For relationships, ambiguity is mainly in polyphonic scores, where
annotators had to decide how to attach noteheads from multiple
voices to crescendo and decrescendo hairpin symbols.
However, after quality control, there were 689 -- 691 objects in the image
and 613 -- 637 relationships, depending on which annotator we asked.
This highlights the limits of both the annotation guidelines and QC:
the ground truth is probably not entirely unambiguous, so various
configurations passed QC, and additionally the QC process itself allows 
for human error. (If we could really automate infallible QC, 
we would also have solved OMR!)

At the same time, as seen in table \ref{tab:data:agreement},
the two-round quality control process
apparently removed nearly four fifths of all disagreements, bringing the withQC 
inter-annotator f-score of 0.97 from a noQC f-score of 0.89. On average,
quality control introduced {\em less} change than was originally between
individual annotators. This statistic seems to suggest that the withQC results are 
somewhere in the ``center'' of the space of submitted
annotations, and therefore the quality control process really leads to more
accurate annotation instead of merely distorting the results in its own way.

We can conclude that the annotations, using the quality control process,
is quite reliable, even though slight mistakes may remain. Overfitting the test set
will likely not be an issue.

\subsection{Known limitations}
\label{subsec:data:limitations}

The MUSCIMA++ 1.0 dataset is far from perfect, as is always the case with extensive
human-annotated datasets. In the interest of full disclosure and managing
expectations, we list the known issues.

Annotators also might have made mistakes that slipped both through automated
validation and manual quality control. In automated validation, there is
a tradeoff between catching errors and false alarms: 
events like multiple stems per notehead happen even
in the limited set of 20 pages of MUSCIMA++. 
In the same vein, although we did implement automated checks for 
highly inaccurate annotations, they only catch some
of the problems as well, and our manual quality control procedure also relies
on inherently imperfect human judgment. All in all, the data is not perfect.
With limited man-hours, there is always a tradeoff between quality and scale.

The CVC-MUSCIMA dataset has had staff lines removed automatically with very high
accuracy, based on a precise writing and scanning setup (using a standard notation
paper and a specific pen across all 50 writers). However, there are still some
errors in staff removal: sometimes, the staff removal algorithm took with it
some pixels that were also legitimate part of a symbol. This manifests itself
most frequently with stems.

The relationship model is rather basic. Precedence and simultaneity relationships
are not annotated, and stafflines and staves are not among the annotated
symbols, so notehead-to-staff assignment is not explicit. Similarly,
notehead-to-measure assignment is also not explicitly marked.
This is a limitation that so far does not enable inferring pitch from the
ground truth. However, much of this should be obtainable automatically,
from the annotation that is available and the CVC-MUSCIMA staff removal
ground truth.

There are also some outstanding technical issues in the details of how
the bridge from graphical expression to interpretation ground truth is designed.
For example,
there is no good way to encode a 12/8 time signature. The "1" and "2" 
would currently be annotated as separate numerals, and the fact that
they belong together to encode the number 12 is not represented explicitly:
one would have to infer that from knowledge of time signatures. A potential
fix is to introduce a ``numeric\_text`` symbol as an intermediary between
``time\_signature`` and ``numeral\_X`` symbols, similarly to various
``(some)\_text`` symbols that group ``letter\_X`` symbols.
Another technical problem is that the mask of empty noteheads that 
lie on a ledger line includes the part of the ledger line that lies
within the notehead.

Finally, there is no good policy on symbols broken into two at line breaks.
They are currently handled as two separate symbols.

\section{Conclusion}
\label{sec:conclusion}


In MUSCIMA++ v.\currentversion, we provide an OMR dataset of handwritten
music that allows training and benchmarking OMR systems tackling
the symbol recognition and notation reconstruction stages of the OMR
pipeline. Building on the CVC-MUSCIMA staff removal ground truth,
we provide ground truth for symbol localization, classification,
and symbol graph recovery, which is the step that resolves ambiguities
necessary for inferring pitch and duration. Although the dataset does
not explicitly record precedence, simultaneity, and attachment to stafflines
and staves, this information can be inferred automatically from the staff
removal ground truth from CVC-MUSCIMA and the existing MUSCIMA++
symbol annotations.

\subsection{What can we do with MUSCIMA++?}
\label{subsec:conclusion:tasks}

MUSCIMA++ allows evaluating OMR performance on various sub-tasks in isolation.

\begin{itemize}[noitemsep]
\item {\bf Symbol classification:} use the bounding boxes and symbol masks as inputs, symbol labels as outputs. Use primitive relationships to generate a ground truth of composite symbols, for compatibility with datasets of \cite{CalvoZaragoza2014} or \cite{Rebelo2010}.
\item {\bf Symbol localization:} use the pages (or sub-regions) as inputs; the corresponding list of bounding boxes (and optionally, masks) is the output.w
\item {\bf Primitives assembly:} use the bounding boxes/masks and labels as inputs, adjacency matrix as output. (For MUSCIMA++ version~\currentversion,
be aware of the limitations discussed in \ref{subsec:data:limitations}.)
\end{itemize}

\noindent
At the same time, these inputs and outputs can be chained, to evaluate
systems tackling these sub-tasks jointly.

What about inferring pitch and duration?

First of all, we can exploit the 1:1 correspondence between notes and noteheads.
Pitch and duration can therefore be thought of as extra attributes of notehead-class
symbols.
Duration of notes (at least, relative -- half, quarter, etc.) can then already be
extracted from the annotated relationships of MUSCIMA++~\currentversion.
Tuplet symbols
are also linked explicitly to the notes (noteheads) they affect.
To reconstruct pitch, though, one needs to go beyond what
MUSCIMA++~\currentversion~makes explicitly available. 
The missing elements of ground truth
are relationships that attach key signatures and noteheads to stafflines, and
secondarily to measures (the scope of ``inline'' accidentals is until the next bar).
However, these relationships should be relatively straightforward to add
automatically, using the CVC-MUSICMA staffline ground truth.

\subsection{Future work}
\label{subsec:conclusion:futurework}

MUSCIMA++ in its curent version~\currentversion~does not
entirely live up to the
requirements discussed in \ref{sec:groundtruth} and \ref{sec:choiceofdata}
-- several areas need improvement.

First of all, the relationship model is so far rather basic. While
it should be possible to automatically infer precedence and simultaneity,
stafflines, staves, and the relationships of noteheads to the staff
symbols, it is not entirely clear how accurately it can be done.
This is the major obstacle to automatically inferring pitch
from the currently available data.

Second, the source of the data is relatively limited, to the collection
effort of CVC-MUSCIMA. While the variety of handwriting collected by
Forn\'{e}s et al. \cite{Fornes2012} is impressive,
it is all {\em contemporary} -- whereas the application domain of handwritten
OMR is also in early music, where different handwriting styles have been used.
The dataset should be enriched by early music sources.

Third, the current {\em ad hoc} data format is not a good long-term solution.
While it encodes the ground truth information well, and we do provide tools
for parsing, visualization and implement a data model, it would be beneficial
to re-encode the data using MEI. This does {\em not} mean only using the
graphics-independent MEI XML format for recording the musical content -- MEI
also has the capacity to encode the graphical elements, their locations,
the input pages, etc. For relationships, MEI allows user-defined elements.
Re-encoding the dataset in MEI would also lift the burden 
of maintaining an independent data model implementation.

Finally, now that MUSCIMA++ is available, there is sufficient data to train
models for automating annotation. This could significantly speed up future work
and enable us to cover a greater variety of scores, including those from the existing
dataset of Rebelo et al. \cite{Rebelo2012Thesis}. The resources freed up
through automation could also be used to annotate staff removal ground truth
in other scores, which by itself takes anecdotally about as much time as annotating all
the remaining symbols.

\subsection{Final remarks}
\label{subsec:conslusions:finalremarks}

In spite of its imperfections, the MUSCIMA++ dataset
still offers the most complete and extensive publicly available ground truth annotation
for OMR to date. Together with the provided software, it should
enable the OMR field to establish a robust basis for comparing systems
and measuring progress. Organizing a competition, as called for
in \cite{Byrd2015}, would be a logical next step towards this end. Although
specific evaluation procedures will need to be developed for this data,
we believe the fine-grained annotation will enable evaluating at least the stage 2
and stage 3 tasks in isolation and jointly, with a methodology 
analogous to those suggested in \cite{Byrd2015}, \cite{Droettboom2004}, 
or \cite{Bellini2007}. 
Finally, it can also serve as the training data for extending
the machine learning paradigm of OMR described 
by Calvo-Zaragoza et al. \cite{CalvoZaragoza2017}
to symbol recognition and notation assembly tasks.

Our hope is that the MUSCIMA++ dataset will be useful 
to the broad OMR community.

\section*{Acknowledgment}

First of all, we thank our annotators, who did much of the tedious
work. 
The authors would also like to express their gratitude to Alicia Forn\'{e}s
of CVC UAB\footnote{\url{http://www.cvc.uab.es/people/afornes/}}
who generously decided to share the CVC-MUSICMA dataset
under the CC-BY-NC-SA 4.0 license, thus enabling us to share
the MUSCIMA++ dataset in the same open manner as well.

This work is supported by the Czech Science Foundation,
grant number P103/12/G084.



\bibliographystyle{IEEEtran}
\bibliography{bibliography}
%
%
%

\end{document}